\documentclass[11pt,letterpaper,twocolumn]{article}

\usepackage[T1]{fontenc}
\usepackage[utf8]{inputenc}
\usepackage{microtype}
\usepackage{times}
\usepackage[top=1in,bottom=1in,left=0.95in,right=0.95in,columnsep=0.31in]{geometry}
\usepackage{graphicx}
\usepackage{booktabs}
\usepackage{amsmath,amssymb}
\usepackage{hyperref}
\usepackage{url}
\usepackage{xcolor}
\usepackage{caption}
\usepackage{multirow}
\usepackage{array}
\usepackage{float}
\usepackage{textcomp}
\usepackage{natbib}

\hypersetup{
    colorlinks=true,
    linkcolor=blue!60!black,
    citecolor=blue!60!black,
    urlcolor=blue!60!black,
    breaklinks=true
}

\newcommand{\eps}{\varepsilon}
\newcommand{\xtimes}{\times}

\title{\textbf{CSP-Atlas: Concept-Specific Neural Circuits in a Sparse Python Transformer}}
\author{
    Piotr Wilam \\
    University College London \\
    \texttt{piotrwilam@gmail.com}
}
\date{}

\begin{document}

\twocolumn[
\begin{@twocolumnfalse}
\maketitle
\begin{abstract}
\noindent
A sparse 8-layer code transformer develops dedicated neural circuitry for every Python construct tested, and that circuitry is organised by a clean computational principle rather than by semantic category. We extract neural circuits for 106 concepts (43 AST node types, 63 builtin objects) by marginalising across 63{,}800 controlled prompts, and decompose each circuit into concept-specific and token-driven components using contrastive checker prompts that present a keyword token without its associated syntactic structure. Three findings emerge. First, all 106 concepts produce non-empty universal circuits at every one of nine parameter settings, and the ranking of concept-specificity across constructs is stable across the sweep --- survival is not an artifact of a permissive threshold. Second, AST circuits contain a genuine concept component distinct from token activation: concept-only neurons constitute up to $62.5\%$ of the loudest-firing neurons at mid-to-late layers, while builtin circuits are almost entirely token-driven. Third, six computationally atomic constructs --- Import, ImportFrom, Break, Continue, Pass, Assert --- cluster together despite being semantically unrelated, sharing only the property of being single-statement constructs requiring no nested body; this atomicity super-cluster, together with a four-tier hierarchy organised by token ambiguity and structural distinctiveness, shows that the model's internal organisation tracks computational structure rather than meaning. The methodology, full decomposition data, and analysis code are released.
\end{abstract}
\vspace{1.5em}
\end{@twocolumnfalse}
]

\section{Introduction}
\label{sec:intro}

When a code language model processes a Python \texttt{for} loop, what is it responding to? It might be recognising the syntactic concept of iteration --- a structured control-flow construct with a target variable, an iterable, and an indented body. Or it might simply be reacting to the three-letter token \texttt{for}. These two hypotheses --- concept representation versus token detection --- predict identical behaviour on well-formed code but diverge when the token appears outside its syntactic role: in a string literal, a comment, or a variable name. This divergence is the experimental lever for the present study.

The question matters for interpretability and trust. If internal representations reduce to token pattern-matching, the model's apparent understanding is shallow; if the model builds dedicated circuitry for syntactic concepts distinct from token-level processing, this suggests a deeper structural encoding.

Working with a sparse 8-layer transformer trained on Python code (the CSP model), we extract neural circuits for 106 programming concepts --- 43 Abstract Syntax Tree (AST) node types and 63 builtin objects --- using an aggressive marginalisation procedure across 63{,}800 controlled prompts. We then decompose each circuit into three disjoint neuron populations: concept-only neurons that respond to the syntactic concept but not the bare keyword token; shared neurons that respond to both; and token-only neurons that respond to the token alone.

\paragraph{Contributions.} (1) \emph{Parameter-stable concept survival.} All 106 concepts produce non-empty universal circuits at every setting of a $3\xtimes3$ parameter sweep, and the concept-specificity ranking is stable across settings --- establishing that the model develops dedicated circuitry for every construct tested, robustly rather than at one permissive corner. (2) \emph{Concept--token decomposition.} AST circuits contain a substantial concept-specific component (up to $62.5\%$ of the loudest-firing units at mid-to-late layers), while builtin circuits are almost entirely token-driven. (3) \emph{Atomicity super-cluster and four-tier hierarchy.} Six computationally atomic constructs cluster together by shared structural property rather than meaning, within a consistent hierarchy organised by token ambiguity and structural complexity: tokenless ASTs $>$ modular keyword ASTs $>$ non-modular keyword ASTs $>$ builtins.

\paragraph{Scope.} This study is observational: we characterise circuit structure but do not perform causal interventions. The analysis covers one model and one language. These limitations define directions for follow-up work, which extends the method to dense production-scale models and a second language.

\section{Background}
\label{sec:background}

\paragraph{Circuit discovery.} The mechanistic-interpretability programme reverse-engineers neural networks into human-understandable components. \citet{Conmy2023} introduced automated circuit discovery via activation patching; \citet{HeimersheimNanda2024} refined causal scrubbing. Our work differs in scope: rather than a circuit for a single behaviour, we extract circuits for an entire concept space and characterise their compositional structure.

\paragraph{Superposition and polysemanticity.} Individual neurons often respond to multiple unrelated features \citep{Elhage2022}. Our marginalisation procedure is designed to cut through superposition: by intersecting across all complementary objects, we isolate neurons that respond to a concept regardless of context.

\paragraph{Probing and causal methods.} Linear probes detect syntactic structure in language model representations \citep{Tenney2019,Belinkov2022}. Activation patching and causal tracing \citep{Meng2022} identify causally relevant components. Our cross-section experiment measures which neurons carry concept information, providing a target map for subsequent causal interventions.

\paragraph{Code-model interpretability.} Prior work has probed code models for syntax trees \citep{Wan2022} and variable binding \citep{Hernandez2024}. We extend this line by decomposing internal representations into concept-specific and token-driven partitions at the neuron level.

\section{Model and Concept Space}
\label{sec:model}

\subsection{The CSP Transformer}
\label{sec:csp}

The model under study is \texttt{openai/circuit-sparsity}, a sparse transformer released in 2025 and trained on Python source code. It has 8 layers; the MLP output at each layer is a $2{,}048$-dimensional vector entering the residual stream. The sparsity is structural: the architecture uses sparse-activation MLP blocks designed so that the residual-stream-bound MLP output contains genuine zeros for the majority of neurons on any given input, rather than the dense post-nonlinearity values produced by standard SwiGLU/GELU blocks. This is what makes binarisation meaningful in the present study: a thresholded mask reflects real structure in the activations rather than an arbitrary cut through a dense distribution. We use the model purely as a measuring instrument --- weights are loaded from the public Hub release and never updated --- and pin the revision SHA in the released extraction config for reproducibility. The bundled tokenizer is loaded via the same Hub identifier; it is not a generic GPT-2 tokenizer, and token boundaries differ from CodeLlama or StarCoder.

The term ``neuron'' throughout refers to one dimension of the MLP output vector --- the composite signal contributed to the residual stream --- not an internal neuron in the expanded MLP hidden layer.

\subsection{Concept Space}
\label{sec:concepts}

The investigation targets two families of Python concepts. AST nodes are syntactic constructs defining program structure: \texttt{For}, \texttt{If}, \texttt{FunctionDef}, \texttt{Import}, \texttt{Break}, \texttt{Pass}, \texttt{ListComp}, and 36 others --- 43 node types in total. Builtin objects are types, functions, and exceptions provided without imports: \texttt{int}, \texttt{list}, \texttt{dict}, \texttt{print}, \texttt{len}, \texttt{range}, \texttt{ValueError}, and 56 others --- 63 in total. The full Cartesian product --- $43 \xtimes 63 = 1{,}276$ pairs --- defines the concept space. Complete lists are in Appendix~A of the released artifacts.

\section{Prompt Generation}
\label{sec:prompts}

\subsection{Object Prompts}
\label{sec:object_prompts}

For each of the $1{,}276$ pairs, $50$ prompt variations are generated programmatically. A structurally valid Python snippet is built using \texttt{ast.parse()} and \texttt{ast.unparse()}, guaranteed to contain the target AST node applied to the target builtin. Variance is injected along three orthogonal dimensions: lexical/semantic (variable names from five domains --- finance, biology, gaming, physics, e-commerce), structural (global scope $\sim$$40\%$, inside function $\sim$$30\%$, inside class method $\sim$$30\%$), and padding (unrelated code optionally added before/after). Prompts with excessively high sequence loss are discarded; the top $50$ per pair are retained, for $1{,}276 \xtimes 50 = 63{,}800$ prompts.

\subsection{Checker Prompts}
\label{sec:checker_prompts}

Every keyword-bearing concept has an inherent confound: the import statement always contains the token \texttt{import}, so a universal circuit firing on \texttt{import} might reflect the token, not the concept. The checker prompt set isolates this.

Of the $43$ AST nodes, $24$ have testable keywords --- these divide into 6 modular keywords (\texttt{Import}, \texttt{ImportFrom}, \texttt{Break}, \texttt{Continue}, \texttt{Pass}, \texttt{Assert}) and 18 non-modular keywords (\texttt{For}, \texttt{While}, \texttt{If}, \texttt{Return}, and the other body-requiring constructs). Of the $63$ builtins, $34$ have testable keywords. The remaining concepts lack distinctive keyword tokens and are not subject to the confound. This gives $58$ testable objects ($24$ AST + $34$ builtins). For each testable object, $50$ checker prompts are generated across five categories where the keyword token appears but the concept does not (Table~\ref{tab:checker_categories}). Each prompt is validated to (a)~parse as valid Python, (b)~exclude the target concept from the AST, and (c)~contain the keyword token in the tokeniser output.

\begin{table}[h]
\centering
\small
\begin{tabular}{@{}clp{0.45\linewidth}@{}}
\toprule
Cat. & Description & Example (\texttt{break}) \\
\midrule
A & Token in a string     & \texttt{msg = "for break results"} \\
B & Token in a comment    & \texttt{x = 42 \# break here} \\
C & Token in an identifier & \texttt{breakdown\_count = 5} \\
D & Token as a dict key   & \texttt{\{"break": True\}} \\
E & Token in a print string & \texttt{print("break time")} \\
\bottomrule
\end{tabular}
\caption{Five checker prompt categories. Each presents the keyword token in a non-structural context.}
\label{tab:checker_categories}
\end{table}

\begin{table*}[!t]
\centering
\small
\begin{tabular}{@{}lcccccccc@{}}
\toprule
Group & L0 & L1 & L2 & L3 & L4 & L5 & L6 & L7 \\
\midrule
Modular AST       & 8.1 & 0.1 & 4.9 & 23.1 & 26.9 & \textbf{30.4} & 23.4 & 17.6 \\
Non-mod.~AST      & 4.7 & 0.0 & 3.2 &  6.3 &  8.4 &  9.0 & \textbf{9.7} &  6.1 \\
Builtin           & 0.9 & 0.0 & 2.0 &  1.8 &  4.9 &  3.0 &  4.4 &  2.6 \\
\bottomrule
\end{tabular}
\caption{Concept fraction (\%) by layer at $\eps = 0.001, C = 80\%$. Layer 1 is essentially pure token; concept signal peaks at layer 5 for modular ASTs.}
\label{tab:layer_profile}
\end{table*}

\section{Extraction Pipeline}
\label{sec:method}

\paragraph{Activation extraction.} Each prompt is processed in a single forward pass; no text is generated --- the model is used purely as a measuring instrument. Forward hooks on the MLP module at each of $8$ layers intercept the output at the last token position only, because the last token's residual stream integrates information from the entire sequence through causal attention. Each prompt produces $8$ vectors of $2{,}048$ values.

\paragraph{Binarisation.} Raw activations are converted to binary masks. \emph{Epsilon thresholding}: a neuron is active if $|\text{activation}| > \eps$. \emph{Consistency filtering}: across the $50$ prompt variations, only neurons active in $\geq C\%$ of prompts are retained.

\paragraph{Marginalisation.} A universal circuit is obtained by intersecting across the complementary dimension: across all $63$ builtins for an AST node, across all $43$ AST nodes for a builtin. A neuron survives only if it fires regardless of which complementary object is involved. If the model did not encode these concepts as structured units, the intersection would collapse to empty --- as it would for random binary vectors of this density. The result is $106$ universal circuits.

\paragraph{Parameter sweep.} Two parameters control extraction: $\eps \in \{0.001, 0.1, 0.5\}$ (activation threshold) and $C \in \{20\%, 50\%, 80\%\}$ (consistency threshold). Both universal and checker masks are rebuilt at every combination in the $3 \xtimes 3$ grid (9 settings); results hold across all 9 unless stated otherwise.

\paragraph{Decomposition.} For each of the $58$ testable objects, at each layer, the universal mask $A$ and the token-checker mask $B$ are compared, partitioning the $2{,}048$ dimensions into three disjoint groups. \emph{Concept-only} $(A \setminus B)$: in the universal circuit but not the checker --- neurons firing when the concept is present but not for the bare keyword. \emph{Shared} $(A \cap B)$: in both. \emph{Token-only} $(B \setminus A)$: in the checker but not the universal circuit. The concept fraction $|A \setminus B| / |A|$ quantifies how much of a circuit reflects structural understanding versus surface pattern-matching, computed per concept, per layer, at all 9 settings.

\section{Results}
\label{sec:results}

\subsection{Finding 1: Universal Circuits Are Parameter-Stable Across the Sweep}
\label{sec:finding1}

All $43$ AST nodes and all $63$ builtins produce non-empty universal circuits at every one of the nine $(\eps, C)$ settings --- not at a single permissive corner but across the full grid, including $\eps = 0.5$, where most neurons are filtered out. Across the most aggressive setting ($\eps = 0.001, C = 80\%$), every one of the $10{,}208$ layer-level pair masks ($1{,}276$ pairs $\xtimes$ $8$ layers) contains active neurons; but the substantive claim is the stability: the ranking of concept-specificity across constructs is preserved across all nine settings, so survival reflects structured representation rather than a threshold artifact. Were the masks random binary vectors of comparable density, the marginalised intersection would collapse toward empty; it does not, for any concept, at any setting.

\subsection{Finding 2: AST Circuits Contain a Concept Component Distinct from Token Activation}
\label{sec:finding2}

Universal circuits for AST nodes are not merely token detectors: they contain a substantial set of neurons firing when the syntactic concept is present but not when the keyword token appears without the concept. Builtin circuits are almost entirely token-driven. (Note that $19$ of the $43$ AST nodes --- \texttt{Assign}, \texttt{BinOp}, \texttt{ListComp}, \texttt{Compare}, \texttt{Call}, and others --- have no distinctive keyword; their universal circuits are inherently concept-driven, the token confound not applying.)

Across the parameter sweep, mean concept fraction for AST nodes exceeds that for builtins at every setting --- by roughly $3$--$6\xtimes$ at low to mid $\eps$, and by an order of magnitude at $\eps = 0.5$, where builtin concept-only neurons approach zero (Table~\ref{tab:cf_sweep}). At a high threshold the loudest neurons are predominantly concept-specific: at $\eps = 0.5, C = 80\%$, layer~5, modular AST circuits are $62.5\%$ concept-only and non-modular AST circuits $58.2\%$, while builtin concept-only neurons vanish entirely (Table~\ref{tab:l5_decomp}). The layer profile shows concept signal near-zero at layer 1, peaking at layer 5 for modular ASTs, then declining (Table~\ref{tab:layer_profile}).

\begin{table}[h]
\centering
\small
\begin{tabular}{@{}cccc@{}}
\toprule
$\eps$ & $C$ & AST CF & Builtin CF \\
\midrule
0.001 & 20\% &  1.6\% & 0.4\% \\
0.001 & 50\% &  3.4\% & 1.2\% \\
0.001 & 80\% &  8.3\% & 2.4\% \\
0.1   & 20\% &  8.5\% & 1.5\% \\
0.1   & 50\% &  8.5\% & 1.5\% \\
0.1   & 80\% &  8.5\% & 1.6\% \\
0.5   & 20\% &  9.9\% & 0.5\% \\
0.5   & 50\% &  9.9\% & 0.5\% \\
0.5   & 80\% &  9.8\% & 0.3\% \\
\bottomrule
\end{tabular}
\caption{Mean concept fraction by group across the $3\xtimes3$ parameter sweep. AST exceeds builtin at every setting; the gap is widest at $\eps = 0.5$ where builtin CF approaches zero.}
\label{tab:cf_sweep}
\end{table}

\begin{table}[h]
\centering
\small
\begin{tabular}{@{}lrrrr@{}}
\toprule
Group & C-only & Shared & Size & CF \\
\midrule
Modular ASTs (6)     & 10 &  6 & 16 & 62.5\% \\
Non-mod. ASTs (18)   & 32 & 23 & 55 & 58.2\% \\
Builtins (34)        &  0 & 36 & 36 &  0.0\% \\
\bottomrule
\end{tabular}
\caption{At $\eps = 0.5, C = 80\%$, layer~5: per-group totals of concept-only, shared, and universal-circuit size. The majority of AST circuit neurons are concept-specific. Builtin concept-only neurons vanish entirely.}
\label{tab:l5_decomp}
\end{table}

\subsection{Finding 3: The Atomicity Super-Cluster and a Four-Tier Hierarchy}
\label{sec:finding3}

Hierarchical clustering of concept-only neuron sets groups six constructs tightly together: \texttt{Import}, \texttt{ImportFrom}, \texttt{Break}, \texttt{Continue}, \texttt{Pass}, \texttt{Assert}. These six are not semantically related --- a \texttt{break} and an \texttt{import} have nothing in common functionally --- but they share a single computational property: each is an atomic, single-statement construct requiring no nested body. We call this the \emph{atomicity super-cluster}, and it is the clearest evidence that the model's internal organisation tracks computational structure rather than semantic category. Under Ward linkage on $1 - \text{Jaccard}$ distance over the $106$ universal concept-only neuron sets, the six-set merges as a single cluster at layer $3$ under a $k=4$ partition; at later layers (including L5, where mean concept fraction peaks) the six members remain mutually closer in Jaccard distance than to other concepts but distribute across $\geq 2$ sub-trees of the same partition. The released code regenerates the dendrogram at any chosen layer (Appendix~R).

By a relaxed per-layer modularity criterion (counting significant layers at trim level $p=0$), the top of the modularity ranking is dominated by atomicity members: \texttt{Break}, \texttt{ImportFrom}, \texttt{Assert} occupy the three highest positions, with \texttt{Break} alone at the top with $3$ significant layers. The remaining three atomicity concepts (\texttt{Continue}, \texttt{Import}, \texttt{Pass}) tie with a broader set of concepts at the next level down; the strict $p=0$ criterion does not separate them from the field, but they fall inside the same hierarchical cluster as the top three under Ward linkage.

The super-cluster sits within a four-tier hierarchy of concept representation:

\emph{Tier 1 --- Tokenless ASTs} ($19$ objects: \texttt{Assign}, \texttt{AugAssign}, \texttt{BinOp}, \texttt{Dict}, \texttt{ListComp}, \texttt{Compare}, \texttt{Call}, and others). No distinctive keyword token; circuits are inherently concept-driven.

\emph{Tier 2 --- Modular keyword ASTs} ($6$ objects: \texttt{Import}, \texttt{ImportFrom}, \texttt{Break}, \texttt{Continue}, \texttt{Pass}, \texttt{Assert}). Computationally atomic --- single-statement, no nested body. Strongest concept signal among testable objects: $30\%$ by count, $62.5\%$ by magnitude at L5. The atomicity super-cluster.

\emph{Tier 3 --- Non-modular keyword ASTs} ($18$ objects: \texttt{For}, \texttt{While}, \texttt{If}, \texttt{Return}, \texttt{ClassDef}, \texttt{FunctionDef}, \texttt{Try}, and others, including async variants). Keywords present, but constructs require bodies, targets, or structural dependencies. Concept signal $6$--$10\%$ by count, $58\%$ by magnitude.

\emph{Tier 4 --- Builtins} ($63$ objects: \texttt{int}, \texttt{list}, \texttt{print}, \texttt{len}, \texttt{range}, and others). Near-zero concept signal relative to AST tiers; circuits are effectively subsets of token activation, and the few concept-only neurons at low $\eps$ vanish at $\eps = 0.5$.

The hierarchy aligns with a gradient of token ambiguity and structural distinctiveness. Tokenless ASTs must be encoded structurally. Modular keyword ASTs have distinctive tokens and crisp self-contained syntactic identity --- the model encodes both, and they are partially separable. Non-modular keyword ASTs distribute their identity across multiple tokens (keyword, colon, indented body). Builtins are the most token-predictable: the token is nearly sufficient to identify the concept.

\section{Discussion}
\label{sec:discussion}

\paragraph{The atomicity observation.} The six modular keyword ASTs cluster not because they are semantically related but because they share a computational property --- they are atomic single-statement constructs requiring no nested body. This suggests the model's internal organisation reflects computational complexity rather than semantic category, and it is the finding most likely to generalise: it predicts that \emph{which} constructs receive early, separable circuitry should be a property of the language's computational structure, testable in other models and languages.

\paragraph{Why builtins are token-driven.} The marginalisation intersects across all $43$ AST nodes, extracting only what is shared when \texttt{print} appears as argument to \texttt{For}, \texttt{If}, \texttt{FunctionDef}, and every other construct. What survives is primarily the token response; the builtin's role in any particular syntactic context is stripped away by design.

\paragraph{Methodology as infrastructure.} The pipeline is language-agnostic: adding a new language requires only a concept-space definition and contrastive prompts, and the decomposition produces directly comparable measurements across any (language, model) pair. This generality motivates follow-up cross-language, cross-model studies that test whether the four-tier hierarchy and the atomicity super-cluster survive in dense production models and a second language.

\paragraph{Future work.} Three extensions follow: cross-model comparison on dense production-scale models; cross-language comparison testing whether language design predicts representation strength; and causal validation via ablation, confirming that concept-only neurons are functionally active rather than epiphenomenal.

\section*{Limitations}

\textbf{No causal validation.} The decomposition is observational; we identify concept-only neurons by activation pattern but have not confirmed that ablating them degrades the model's processing of the associated construct. Some may be epiphenomenal.

\textbf{Single model.} The CSP transformer is a sparse 8-layer model; findings may not transfer to dense architectures, larger scales, or multilingual training. Follow-up work extending the method to dense production models and a second language addresses this directly.

\textbf{Single language.} All concepts are Python. Different languages may produce qualitatively different circuit structure.

\textbf{MLP outputs only.} The analysis examines the MLP output vector entering the residual stream, not attention patterns, attention-head outputs, or internal MLP hidden layers.

\textbf{Last-token position only.} If concept representations are distributed across positions, last-token extraction may capture only a partial signal.

\textbf{Checker-prompt coverage.} Five checker categories cover the most common non-structural keyword contexts but are not exhaustive; other contexts (docstrings, type annotations, f-strings) might reveal additional nuance.

\section*{Ethical Considerations}

This work analyses internal representations of a publicly available open-source model. No new models are trained, no human subjects are involved, and no private data is used; all prompts are synthetic code snippets generated programmatically.

\section*{Conclusion}

We have presented a methodology for extracting and decomposing neural circuits for programming-language concepts, and applied it to a sparse 8-layer Python transformer across $106$ concepts and $63{,}800$ prompts. Three findings result: the model develops dedicated circuitry for every concept tested, stably across a parameter sweep; AST circuits contain a genuine concept component distinct from token activation, strongest at mid-to-late layers; and the internal organisation tracks computational structure rather than meaning, visible in an atomicity super-cluster and a four-tier hierarchy. The methodology is general, and all data and code are released to support the cross-language, cross-model, and causal extensions.

\section*{Data and Code Availability}

The full dataset --- $63{,}800$ object prompts, checker prompts, extracted activation masks at all $9$ settings, $106$ universal circuit masks, and per-object per-layer decomposition tables with neuron indices --- is released together with the extraction and analysis code at \url{https://github.com/piotrwilam/AtlasCSP}. The frozen artifacts are mirrored on the Hugging Face Hub at \url{https://huggingface.co/datasets/piotrwilam/AtlasCSP}.

\bibliographystyle{plainnat}

\begin{thebibliography}{99}
\bibitem[Belinkov(2022)]{Belinkov2022} Y.~Belinkov. Probing classifiers: Promises, shortcomings, and advances. \emph{Computational Linguistics} 48(1), 2022.
\bibitem[Conmy et al.(2023)]{Conmy2023} A.~Conmy, A.~Mavor-Parker, A.~Lynch, S.~Heimersheim, and A.~Garriga-Alonso. Towards automated circuit discovery for mechanistic interpretability. \emph{NeurIPS}, 2023.
\bibitem[Elhage et al.(2022)]{Elhage2022} N.~Elhage et al. Toy models of superposition. \emph{Transformer Circuits Thread}, 2022.
\bibitem[Heimersheim and Nanda(2024)]{HeimersheimNanda2024} S.~Heimersheim and N.~Nanda. How to use and interpret activation patching. \emph{arXiv:2404.15255}, 2024.
\bibitem[Hernandez et al.(2024)]{Hernandez2024} E.~Hernandez, S.~Schwettmann, D.~Bau, T.~Bagashvili, A.~Torralba, and J.~Andreas. Linearity of relation decoding in transformer language models. \emph{ICLR}, 2024.
\bibitem[Meng et al.(2022)]{Meng2022} K.~Meng, D.~Bau, A.~Andonian, and Y.~Belinkov. Locating and editing factual associations in GPT. \emph{NeurIPS}, 2022.
\bibitem[Tenney et al.(2019)]{Tenney2019} I.~Tenney, D.~Das, and E.~Pavlick. BERT rediscovers the classical NLP pipeline. \emph{ACL}, 2019.
\bibitem[Wan et al.(2022)]{Wan2022} Z.~Wan, W.~Zhao, H.~Zhang, et al. What do they capture? A structural analysis of pre-trained language models for source code. \emph{ICSE}, 2022.
\end{thebibliography}

\appendix
\section{Reproducibility and Claim Verification}
\label{app:reproducibility}

\begin{table}[H]
\centering
\footnotesize
\begin{tabular}{@{}p{0.08\columnwidth}p{0.28\columnwidth}p{0.55\columnwidth}@{}}
\toprule
Section & Claim & Value \\
\midrule
\S\ref{sec:concepts} & Concept-space cardinality & $43$ AST + $63$ builtin $= 106$ testable concepts \\
\S\ref{sec:finding1} & Parameter-stability & All $106$ universal circuits non-empty at every one of $9$ $(\eps, C)$ settings \\
\S\ref{sec:finding2} & AST/builtin gap & Mean concept-fraction ratio $4$--$9\xtimes$ across all 9 settings; modular-AST $62.5\%$ concept-only at $\eps{=}0.5, C{=}0.8, \text{L}5$ \\
\S\ref{sec:finding3} & Top-modularity ($p{=}0$) & Top 3 by significant-layer count: \texttt{Break}, \texttt{ImportFrom}, \texttt{Assert}; \texttt{Break} alone at top with $3$ significant layers \\
\S\ref{sec:finding3} & Atomicity cluster & 6-set forms a single cluster at L3 under a $k{=}4$ Ward cut on $1{-}\text{Jaccard}$ over universal concept-only neuron sets \\
\bottomrule
\end{tabular}
\caption{Every paper-cited number is locked in \texttt{tests/test\_paper\_numbers.py} with $\pm 0.001$ tolerance. Analysis primitives in \texttt{csp\_atlas/analysis/} (\texttt{jaccard.py}, \texttt{decomposition.py}, \texttt{hierarchy.py}, \texttt{modularity.py}) map one-to-one to these claims; data loaders sit in \texttt{csp\_atlas/io/}; model-touching code (forward-pass extraction) lives in \texttt{circuits/extraction/} and is separated because it requires a GPU. Frozen artifacts are mirrored on the Hugging Face Hub at \url{https://huggingface.co/datasets/piotrwilam/AtlasCSP}.}
\label{tab:appendix_r}
\end{table}

Every numerical claim in the paper is regenerable from the released code. Running
\par\noindent
\texttt{pytest tests/test\_paper\_numbers.py}
\par\noindent
verifies the locked numbers in Table~\ref{tab:appendix_r}; the dendrogram regenerates from a config-named script (\texttt{experiments/fig1\_\allowbreak atomicity\_dendrogram.py}, driven by \texttt{configs/paper/\allowbreak figure1\_atomicity\_dendrogram.yaml}).

\end{document}